%% file: main.tex
\pdfoutput=1

\documentclass{article}
\usepackage{ijcai19}
\usepackage[T1]{fontenc}
\usepackage{aecompl}
\usepackage{times}
\usepackage{soul}
\usepackage{bbm}

\usepackage{url}
\usepackage[algoruled,linesnumbered]{algorithm2e}
\usepackage[utf8]{inputenc}
\usepackage[small]{caption}
\usepackage{graphicx}
\usepackage{amsmath}
\usepackage{booktabs}
\usepackage{bm}
\usepackage{dsfont}
\usepackage{empheq}
\usepackage{amsfonts}
\usepackage{amssymb}
\usepackage{subfigure}
\usepackage{algorithmic}
\input{commands.tex}

\newcommand\footnoteref[1]{\protected@xdef\@thefnmark{\ref{#1}}\@footnotemark}
\title{Learning Interpretable Deep State Space Model for\\ Probabilistic Time Series Forecasting}
\author{
Longyuan Li$^{1,2}$\and
Junchi Yan$^{2,3}$\footnote{Yaohui Jin and Junchi Yan are corresponding authors. This work is supported by National Key Research and Development Program of China (2018YFC0830400), (2016YFB1001003), STCSM(18DZ1112300).}\and
Xiaokang Yang$^{2,3}$\And
Yaohui Jin$^{1,2*}$\\
\affiliations
$^1$State Key Lab of Advanced Optical Communication System wand Network\\
$^2$MoE Key Lab of Artificial Intelligence, AI
Institute\\
$^3$Department of Computer Science and Engineering\\Shanghai Jiao Tong University\\
\emails
\{jeffli, yanjunchi,xkyang,jinyh\}@sjtu.edu.cn
}

\begin{document}

\maketitle
\begin{abstract}
Probabilistic time series forecasting involves estimating the distribution of future based on its history, which is essential for risk management in downstream decision-making. We propose a deep state space model for probabilistic time series forecasting whereby the non-linear emission model and transition model are parameterized by networks and the dependency is modeled by recurrent neural nets. We take the automatic relevance determination (ARD) view and devise a network to exploit the exogenous variables in addition to time series. In particular, our ARD network can incorporate the uncertainty of the exogenous variables and eventually helps identify useful exogenous variables and suppress those irrelevant for forecasting. The distribution of multi-step ahead forecasts are approximated by Monte Carlo simulation. We show in experiments that our model produces accurate and sharp probabilistic forecasts. The estimated uncertainty of our forecasting also realistically increases over time, in a spontaneous manner.
\end{abstract}

\section{Introduction and Related Works}
\subsection{Background and Motivation}
Time series forecasting is a long-standing problem in literature, which has attracted extensive attention over the decades. For downstream decision-making applications, one important feature for a forecasting model is to issue long-term forecasts with effective uncertainty estimation. Meanwhile the relevant factors, in the form of exogenous variables in addition to the time series data, can be robustly identified, to improve the interpretation of the model and results.

To address these challenges, in this paper, we develop an approach comprising a state space based generative model and a filtering based inference model, whereby the non-linearity of both emission and transition models is achieved by deep networks and so for the dependency over time by recurrent neural nets. Such a network based parameterization provides the flexibility to fit with arbitrary data distribution. The model is also tailored to effectively exploit the exogenous variables along with the time series data, and the multi-step forward forecasting is fulfilled by Monte Carlo simulation. In a nut shell the highlights of our work are:

1) We present a novel deep state space model for probabilistic time series forecasting, which can i) handle arbitrary data distribution via nonlinear network parameterization based on the proposed deep state space model; ii) provide interpretable forecasting by incorporating extra exogenous variable information using the devised ARD network; iii) uncertainty modeling of the exogenous variables to improve the robustness of variable selection. We believe these techniques as a whole are indispensable for a practical forecasting engine and each of them may also be reused in other pipelines.

2) In particular, by taking the automatic relevance determination (ARD) view~\cite{wipf2008new}, we devise an ARD network which can estimate exogenous variables' relevance to the forecasting task. We further consider and devise the Monte Carlo sampling method to model the uncertainty of the exogenous variables. It is expected (and empirically verified) to capture interpretable structures from time series which can also benefit long-term forecast accuracy.
 

3) We conduct experiments on a public real-world benchmark, showing the superiority of our approach against state-of-the-art methods. In particular, we find our approach can spontaneously issue forecasting results with reasonable growing uncertainty over time steps and help uncover the key exogenous variables relevant to the forecasting task.

\subsection{Related Works}
\subsubsection{Time Series Forecasting} Classical models such as auto-regressive (AR) and exponential smoothing \cite{brockwell2002introduction} have a long history for forecasting. They incorporate human priors about time series structure such as trend, seasonality explicitly, and thus have difficulty with diverse dependency and complex structure. With recent development of deep learning, RNNs have been investigated in the context of time series \textit{point} forecasting, the work \cite{langkvist2014review} reviews deep learning methods for time series modelling in various fields. With their capability to learn non-linear dependencies, deep nets can extract high order features from raw time series. Although the results are seemingly encouraging, deterministic RNNs lack the ability to make probabilistic forecasts. \cite{flunkert2017deepar} proposes DeepAR, which employs an auto-regressive RNN with mean and standard deviation as output for probabilistic forecasts.

\subsubsection{Variational Auto-Encoders (VAEs)} Deep generative models are powerful tools for learning representations from complex data \cite{bengio2013representation}. However, posterior inference of non-linear non-Gaussian generative models is commonly recognized as intractable. The idea of variational auto-encoder \cite{kingma2013auto} is to use neural networks as powerful functional approximators to approximate the true posterior, combined with reparameterization tick, learning of deep generative models can be tractable.

\subsubsection{State Space Models (SSMs)} SSMs provide a unified framework for modelling time series data. Classical AR models e.g. ARIMA can be represented in state space form \cite{durbin2012time}. To leverage advances in deep learning, \cite{chung2015recurrent} and \cite{fraccaro2016sequential} draw connections between SSMs and VAEs using an RNN. Deep Kalman filters (DKFs) \cite{krishnan2015deep,krishnan2017structured} further allow exogenous input in the state space models. For forecasts, deep state space models (DSSM) \cite{rangapuram2018deep} use an RNN to generate parameters of a linear-Gaussian state space model (LGSSM) at each time step. Another line of work for building non-linear55ssian assumption of GP-SSMs may be restrictive to non-Gaussian data. Moreover, inference at each time step has at least  o-p;-$O(T)$ complexity for the number of past observations even with sparse GP approximations.

Perhaps the most closely related works are  DSSM \cite{rangapuram2018deep} and DeepAR~\cite{flunkert2017deepar}. Compared with DSSM, the emission model and transition model are non-linear, and our model supports non-Gaussian likelihood. Compared with DeepAR, target values are not used as inputs directly in our approach, making it more robust to noise. Also forecasting samples can be generated computationally more efficiently as the RNN only need to be unrolled once for the entire forecast, whereas DeepAR has to unroll the RNN with entire time series for each sample. Moreover, our model supports exogenous variables with uncertainty as input in the forecasting period. Our model learns relevance of exogenous variables automatically, leading to improvement in interpretability and enhanced forecasting performance.

\subsection{Preliminaries}
\subsubsection{Problem Formulation} Consider we have an $M$-dimensional multi-variate series having $T$ time steps: $\bx_{1:T} = \{ \bx_1, \bx_2,\dots,\bx_T\}$ where each $\bx_t\in \bbR^M$, and $\bx_{1:T} \in \bbR^{M\times T}$. Let $\bu_{1:T+\tau}$ be a set of synchronized time-varying exogenous variables associated with $\bx_{1:T}$, where each $\bu_t\in\bbR^{D}$. Our goal is to estimate the distribution of future window $\bx_{T+1:T+\tau}$ given its history $\bx_{1:T}$ and extended exogenous variables $\bu_{1:T+\tau}$ until the future:
\begin{equation}\label{eq:problem}
    \pT(\bx_{T+1:T+\tau}|\bx_{1:T},\bu_{1:T+\tau})
\end{equation}
where $\bt$ denotes the model parameters. To prevent confusion for \emph{past} and \emph{future}, in the rest of this paper we refer time steps $\{1,2,\dots,T\}$ as \textit{training period}, and $\{T+1,T+2,\dots, T+\tau\}$ as \textit{forecasting period}. The time step $T+1$ is referred to as \textit{forecast start time}, and $\tau\in\mathbb{N}^+$ is the \textit{forecast horizon}. Note that the exogenous variables $\bmu_{1:T+\tau}$ are assumed to be known in both training period and forecasting period, and we allow it to be uncertain and follow some probability distribution $\bu_t\sim p(\bu_t)$ in the forecasting period.

\subsubsection{State Space Models (SSMs)} Given observation sequence $\bx_{1:T}$ and its conditioned exogenous variables $\bu_{1:T}$, a probabilistic latent variable model is:
\begin{equation}
    p(\bx_{1:T}|\bu_{1:T}) = \int p(\bx_{1:T}|\bz_{1:T},\bu_{1:T})p(\bz_{1:T}|\bu_{1:T})\mathrm{d}\bz_{1:T}
\end{equation}\label{eq:LVM}
where $\bz_{1:T}$ for $\bz_t\in\bbR^{n_Z}$ denotes sequence of latent stochastic states. The generative model is composed of an \textit{emission model} $\pT(\bx_{1:T}|\bz_{1:T},\bu_{1:T})$ and a \textit{transition model} $\pT(\bz_{1:T}|\bu_{1:T})$. To obtain state space models, first-order Markov assumption is imposed on the latent sates, and then the emission model and transition model can be factorized:
\begin{equation}\label{eq:ssm_fact}
    \begin{split}
        p(\bx_{1:T}|\bz_{1:T},\bu_{1:T}) &= \prod_{t=1}^{T}p(\bx_t|  \bz_t, \bu_t) \\
        p(\bz_{1:T}|\bu_{1:T}) & = \prod_{t=1}^{T}p(\bz_t|\bz_{t-1},\bu_t)
    \end{split}
\end{equation}
where the initial state w is assumed to be zero. 

\section{Proposed Model}
In this section, we describe i) the deep state space model architecture including both the emission and transition models; ii) the proposed Automatic Relevance Determination (ARD) network to more effectively incorporate the exogenous variable information; iii) the techniques for model training; iv) the devised Monte Carlo simulation for multi-step probabilistic forecasting that can generate seemingly realistic uncertainty estimation over time. We present our approach using a single multi-variate time series sample for notational simplicity. While in fact batches of sequence samples are used for training.
\subsection{Generative Model}\label{sec:GM}
\begin{figure}[tb!]
    \centering
    \subfigure[{The proposed state space based generative model $\pT$}]{\includegraphics[width=0.45\textwidth]{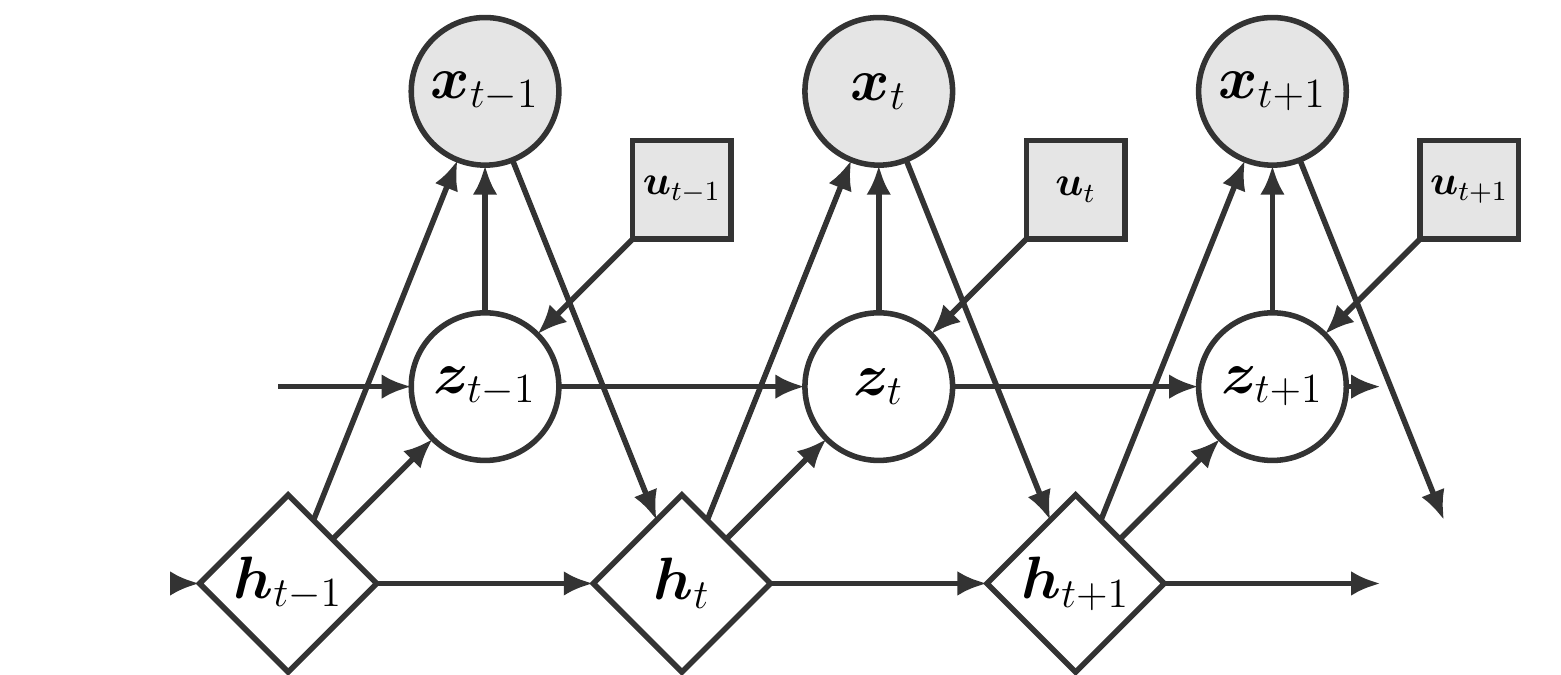}\label{fig:generative}}
    \subfigure[The proposed filtering based inference model $\qPhi$]{\includegraphics[width=0.45\textwidth]{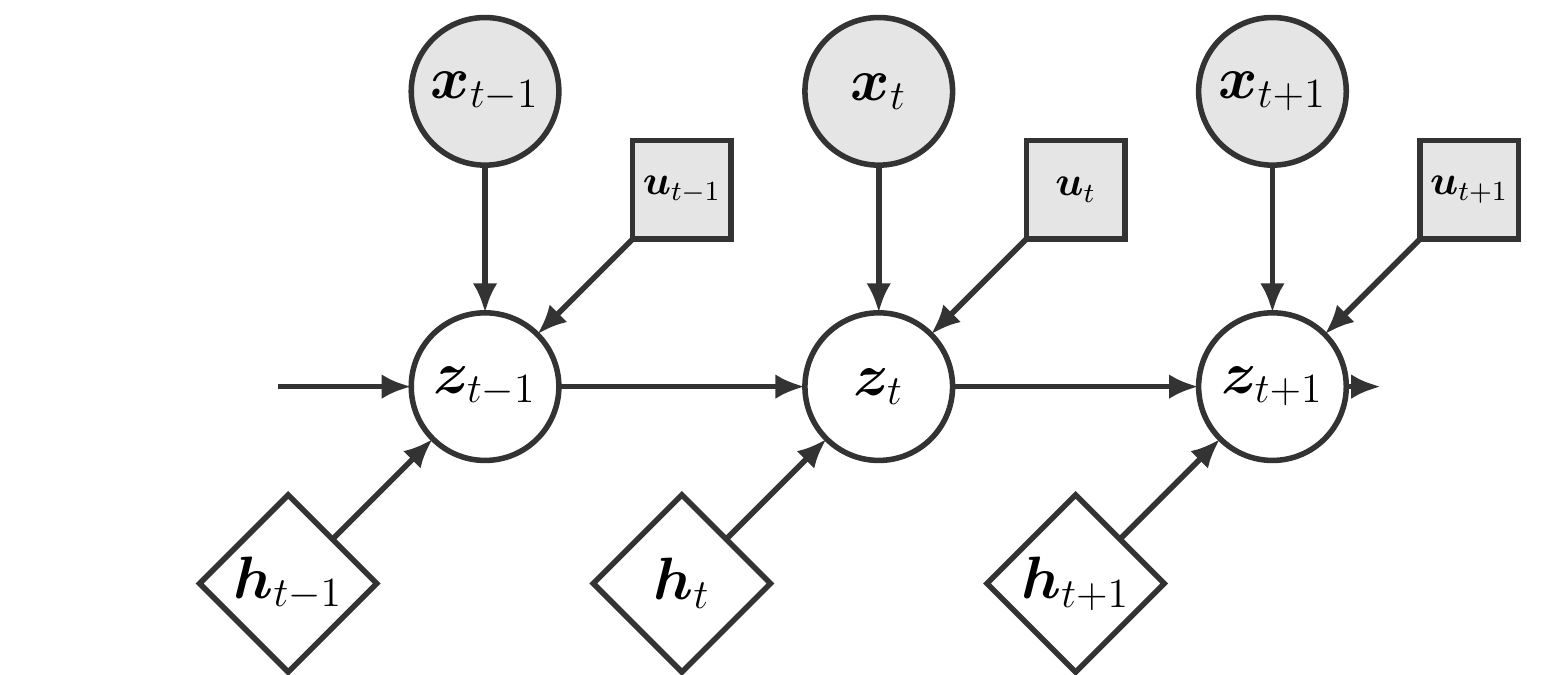}\label{fig:inference}}
    \caption{Proposed generative model and inference model. Shaded and white nodes are observations and hidden variables respectively. Circle and diamond denote stochastic states and deterministic states respectively, and shaded square nodes denote exogenous variables.}
    \label{fig:framework}
\end{figure}
We propose a deep state space model, for the term \emph{deep} it is reflected in two aspects: i) We use neural network to parameterize both the non-linear emission model and transition model; ii) We use recurrent neural networks (RNNs) to capture long-term dependencies.

State space models e.g. linear-Gaussian models and hidden Markov models are neither suitable for modelling long-term dependencies nor non-linear relationships. We relax the first-order Markov assumption in Eq.~\ref{eq:ssm_fact} by including a sequence of deterministic states $\bh_{1:T}$ recursively computed by RNNs to capture the long-term dependencies. Here we simplify the model by assuming the state transition involves no exogenous variable\footnote{In experiments we empirically find using exogenous variables to parameterize the transition model can weaken the performance, which may be due to overfitting and we leave it for future work.}. The generative model is shown in Fig.~\ref{fig:generative}, and the joint probability distribution can be factorized as:
\begin{equation}\label{eq:joint}
\begin{split}
    &\pT(\bx_{1:T}, \bz_{1:T}, \bh_{1:T})\\
    =&\prod_{t=1}^{T}\pT[x](\bx_t|\bz_t,\bh_t,\bu_t)\pT[z](\bz_t|\bz_{t-1},\bh_t)\pT[h](\bh_t|\bh_{t-1},\bx_{t-1})
\end{split}
\end{equation}
We omit initial states $\bz_0=0$ and $\bh_0 = 0$ for brevity, and let $\bx_0 = \bx_1$ for cold starting. In state space form, our model can be represented as:
\begin{align}
    \bh_t&\sim\delta(\bh_{t}-\mathbf{f}_{\bt[d]}(\bh_{t-1},\bx_{t-1}))\label{eq:recurrent}\\
    \bz_t&\sim\mcN(\bmu_{\bt[z]}(\bz_{t-1},\bh_t),\bm{\Sigma}_{\bt{z}}(\bz_{t-1},\bh_t))\label{eq:transition}\\
    \bx_t&\sim\mathcal{PD}(\mathbf{T}_{\bt[x]}(\bz_{t},\bh_t, \bu_t))\label{eq:emission}
\end{align}
where $\mathcal{PD}$ is an arbitrary probability distribution, and its sufficient statistic $\mathbf{T}_{\bt[x]}$ is parameterized by neural networks $\bt[x]$. For stochastic latent states, $\bmu_{\bt[z]}(\cdot)$ and $\bm{\Sigma}_{\bt{z}}(\cdot)$ mean and covariance functions for Gaussian distribution of state transitions, which are also parameterized by neural networks. For deterministic RNN states, $\mathbf{f}_{\bt[d]}(\cdot)$ is the RNN transition function, and $\delta(\cdot)$ is a delta distribution, and  $\bh_t$ can be viewed as following a delta distribution centered at $\mathbf{f}_{\bt[d]}(\bh_{t-1},\bx_{t-1})$.

The deep state space model in Eq. \ref{eq:recurrent},\ref{eq:transition},\ref{eq:emission} is parameterized by $\bt=\{\bt[x], \bt[z], \bt[h]\}$, in our implementation, we use gated recurrent unit (GRU)\cite{cho2014learning} as transition function $\mathbf{f}_{\bt[d]}(\cdot)$ to capture temporal dependency. For stochastic transition, we take the following parameterization:
\begin{equation}
\begin{split}
    \bmu_{\bt[z]}(t) &= \text{NN}_1 (\bz_{t-1}, \bh_t, \bu_t),\\
    \bsigma_{\bt[z]}(t) &= \text{SoftPlus}[\text{NN}_2 (\bz_{t-1}, \bh_t, \bu_t)]
\end{split}
\end{equation}
where $\text{NN}_1$ and $\text{NN}_2$ denotes two neural networks parameterized by $\bt[z]$, and $\text{SoftPlus}[x] = \log(1+\exp(x))$.

For $\mathbf{T}_{\bt[x]}(\bz_{t},\bh_t, \bu_t)$, we also use networks to parameterize the mapping. In order to constrain real-valued output $\tilde{y}=\text{NN}((\bz_{t},\bh_t, \bu_t))$ to the parameter domain, we use the following transformations:
\begin{itemize}
    \item real-valued parameters : no transformation.
    \item positive parameters: the $\text{SoftPlus}$ function.
    \item bounded parameters $[a,b]$: scale and shifted $\text{Sigmoid}$ function $y = (b-a)\frac{1}{1+\exp(-\tilde{y})}+a$
\end{itemize}

\subsection{Structure Inference Network}
We are interested in maximizing the log marginal likelihood, or evidence $\mcL(\bt) = \log \pT(\bx_{1:T}|\bu_{1:T})$, where the latent states and RNN states are integrated out. Integration of deterministic RNN states can be computed by simply substituting the deterministic values. However the stochastic latent states can not be analytically integrated out, because the generative model is non-linear, and moreover the emission distribution may not be conjugate to latent state distribution. We resort to variational inference for approximating intractable posterior distribution \cite{jordan1999introduction}. Instead of maximizing $\mcL(\bt)$, we build a structure inference network $\qPhi$ parameterized by $\bphi$, with the following factorization in Fig.~\ref{fig:inference}:
\begin{equation}\label{eq:qphi_fact}
\begin{split}
    &\qPhi(\bz_{1:T}, \bh_{1:T}|\bx_{1:T},\bu_{1:T}) \\
    =&\prod_{t=1}^{T}\qPhi(\bz_t|\bz_{t-1},\bx_t,\bh_t, \bu_t)\pT[h](\bh_t|\bh_{t-1},\bx_{t-1})
    \end{split}
\end{equation}
where the same RNN transition network structure with the generative model is used. Note that we let stochastic latent states follow a isotropic Gaussian distribution, where covariance matrix $\bm{\Sigma}$ is diagonal: 
\begin{equation}\label{eq:param_inference}
\begin{split}
    \bmu_{\bphi}(t) &= \text{NN}_1 (\bz_{t-1}, \bx_t, \bh_t, \bu_t),\\
    \bsigma_{\bphi}(t) &= \text{SoftPlus}[\text{NN}_2 (\bz_{t-1}, \bx_t, \bh_t, \bu_t)]
\end{split}
\end{equation}

Then we maximize the \textit{variational evidence lower bound (ELBO)} $\mcL(\bt, \bphi)\le\mcL(\bt)$~\cite{jordan1999introduction} with respect to both $\bt$ and $\bphi$, which is given by:
\begin{equation}\label{eq:elbo}
\begin{split}
    \mcL(\bt, \bphi) 
    =& \iint \qPhi\log
    \frac{\pT(\bx_{1:T}, \bz_{1:T}, \bh_{1:T}|\bu_{1:T})}{\qPhi(\bz_{1:T}, \bh_{1:T}|\bx_{1:T},\bu_{1:T})}\mathrm{d}\bz_{1:T}\mathrm{d}\bh_{1:T}\\
    =&\sum_{t=1}^{T}\mathbb{E}_{\qPhi(\bz_t|\bz_{t-1},\bx_t,\bu_t,\bh_t)}\big[\log \pT[x](\bx_t|\bz_t,\bh_t,\bu_t)\big]\\
    &- \text{KL}(\qPhi(\bz_t|\bz_{t-1},\bx_t,\bh_t, \bu_t)\rVert\pT[z](\bz_t|\bz_{t-1},\bh_t))
\end{split}
\end{equation}
where $\qPhi(\bz_{1:T}, \bh_{1:T}|\bx_{1:T},\bu_{1:T})$ is simplified by notation $\qPhi$, and $\text{KL}$ denotes Kullback-Leibler (KL) divergence. Our model jointly learns parameters $\{\bt, \bphi\}$ of the generative model $\pT$ and inference model $\qPhi$ by maximizing ELBO in Eq.~\ref{eq:elbo}. However, latent states $\bz_{1:T}$ are stochastic thus the objective is not differentiable. To obtain differentiable estimation of the ELBO, we resort to Stochastic Gradient Variational Bayes (SGVB): $\mcL_{SGVB}(\bt, \bphi)\simeq\mcL(\bt,\bphi)$, since it provides efficient estimation of ELBO \cite{kingma2013auto}. At each time step, instead of directly sampling from $\qPhi(\bz_t|\bz_{t-1},\bx_t,\bh_t, \bu_t)$, our model samples from an auxiliary random variable $\beps\sim\mcN(\mathbf{0},\mathbf{I})$, and re-parameterizes $\bz_t=\bmu_{\bphi}(t) + \beps\odot\bsigma_{\bphi}(t)$. As such, the gradient of the objective with respect to both $\bt$ and $\bphi$ can be back-propagated through the sampled $\bz_t$. Formally we have:
\begin{equation}
\begin{split}
    \mcL_{\text{SGVB}} = &\sum_{t=1}^{T}\log \pT[x](\bx_t|\bz_t,\bh_t,\bu_t)\\
    &- \text{KL}(\qPhi(\bz_t|\bz_{t-1},\bx_t,\bh_t, \bu_t)\rVert\pT[z](\bz_t|\bz_{t-1},\bh_t))
\end{split}
\end{equation}
The SGVB estimation can be viewed as a single-sample Monte Carlo estimation of the ELBO. To reduce ELBO variance and seek  a tighter bound, we propose to use multi-sampling variational objective \cite{mnih2016variational,burda2015importance}. It can be achieved by sampling $K$ independent ELBOs and taking the average as objective: $\mcL_{\text{SGVB}}^K = \frac{1}{K}\sum_{k=1}^K\mcL_\text{SGVB}^{(k)}$.

\subsection{Automatic Relevance Determination Network}
\begin{figure}[!tb]
    \centering
    \includegraphics[width=0.38\textwidth]{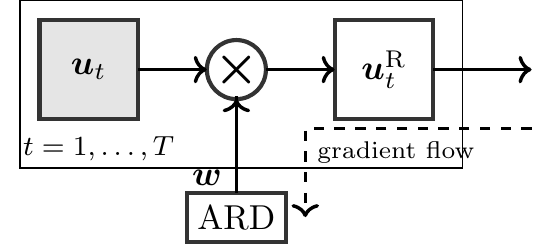}
    \caption{Sketch of the devised automatic relevance determination (ARD) network for modeling associated exogenous variables $\mathbf{u}$.}
    \label{fig:ARD}
\end{figure}

In real-world, in addition to time series it-self, time series data are often correlated to a variety of exogenous variables depending on specific applications, such as: i) time-features: absolute time (trend), hour-of-day, day-of-week, month-of-year (which can carry daily, weekly, yearly seasonality information); ii) flagged periods: such as national holidays, business hours, weekends; iii) measurements: weather, sensor readings. Those features can be incorporated in the form of exogenous variables. In particular, it is often the case that only a few of such variables are relevant to the forecasting task at hand. Existing methods \cite{wen2017multi,rangapuram2018deep,flunkert2017deepar}
select variables based on expert prior. To prune away redundant variables, we take a data-driven sparse Bayesian learning (SPL) approach. Consider given variables $\bu_{1:T+\tau}$, we let:
\begin{equation}
    \bmu_{t}^{\text{R}} = \mathbf{w} \odot \bu_t
\end{equation}
where $\bu_t^{\text{R}}$ are relevant variables, $\mathbf{w}$ is weights of input variables that has the same shape with $\bu_t$. To impose sparsity over variables, we take the so-called automatic relevance determination (ARD)~\cite{neal2012bayesian} view, where the basic idea is to learn an individual weight parameter $\mathbf{w}$ for each input feature to determine relevance. Techniques based on ARD are applied in many classical models, such as SVMs, Ridge regression, Gaussian Processes~\cite{bishop2000variational}. Classical ARD approach is to treat $\mathbf{w}$ as a latent variable placed by some sparsity inducing prior $p(\mathbf{w})$, to avoid complex posterior inference of weights $\mathbf{w}$, we devise an automatic relevant determination (ARD) network whereby $\mathbf{w}$ is parameterized by a neural network with constant input:
\begin{equation}\label{eq:relevance_w}
    \mathbf{w} = \text{SoftMax}(\text{NN}_{\text{ARD}}(\mathbf{I}))
\end{equation}
where $\text{SoftMax}(x_i)=\frac{\exp(x_i)}{\sum_j\exp(x_j)}$, and $\mathbf{I}$ is a constant input vector. We illustrate ARD network in Fig.~\ref{fig:ARD}. Note that $\mathbf{w}$ is a global variable, and multiply every $\bu_t$ for $t=\{1,\dots,T\}$. The differentiable $\text{SoftMax}$ function is used as output layer for multi-class classification as its all outputs sum up to 1. Let $\bu_{1:T}^{R}$ be the input of both generative model and inference model. The gradients of parameters in ARD network can be easily back-propagated within the training process. As such, the ARD network is easy to implement, and it avoids intractable posterior inference of traditional Bayesian approach. We show in experiments that the proposed ARD network not only can learn interpretable structure over time series, but also can improve the accuracy of probabilistic forecasts.

\subsection{Forecasting with Uncertainty Modeling for Exogenous Variables}
Once the parameters of the generative model $\bt$ is learned, we can use the model to address the forecasting problem in Eq.~\ref{eq:problem}. Since the transition model is non-linear, we cannot analytically compute the distribution of future values, so we resort to a Monte Carlo approach.

One shall note that for real-world forecasting, some of the exogenous variables themselves may also be the output by a forecasting model e.g. weather forecast of temperature and rainfall probability. Hence from the forecasting time point, the forward part also contain inherent uncertainty and may not be accurate. Here we adopt a Monte Carlo approach to incorporate such uncertainty into the forecasting process. Let $u_{d,t}\sim p(u_{d,t}), d=1,\dots,D$ be the distribution of each individual exogenous variable $\bu_t\in \bbR^D$ for $t=T+1\dots,T+\tau$ (while those variables with no uncertainty e.g. national holiday can be expressed by a delta distribution), and let $\bu_t\sim p(\bu_t)$ denote the distribution of a collection of exogenous variables at time $t$.

Before forecasting, we first evaluate the whole time series $\bx_{1:T}$ over the model, and compute the distribution of the latent states $\pT[z](\bz_T|\bz_{T-1}, \bh_{T})$ and $\bh_T$ for the last time step $T$ in the training period. Then starting from $\bz_T\sim\pT[z](\bz_T|\bz_{T-1}, \bh_{T})$, we recursively compute:
\begin{equation}\label{eq:mc_forecast}
    \begin{split}
        \bh_{T+t} &= \mathbf{f}_{\bt[d]}(\bh_{T+t-1}, \bx_{T+t-1})\\
        \bz_{T+t} &\sim \mcN(\bmu_{\bt[z]}(\bz_{T+t-1},\bh_{T+t}),\bm{\Sigma}_{\bt{z}}(\bz_{T+t-1},\bh_{T+t}))\\
        \bu_{T+t} &\sim p(\bu_{T+t})\\
        \bx_{T+t} &\sim \mathcal{PD}(\mathbf{T}_{\bt[x]}(\bz_{T+t},\bh_{T+t}, \bu_{T+t}))
    \end{split}
\end{equation}
for $t=1,\dots,\tau$. Our model generates forecasting samples by applying x The sampling procedure can be parallelized easily as sampling recursions are independent.

\subsection{Data Preprocessing}
Given time series $\{\bx_{1},\dots,\bx_{T}\}$, to generate multiple training samples $\{\bx_{1:W}^{i}\}_{i=1}^{N}$, we follow the protocol used in DeepAR \cite{flunkert2017deepar}: the \textit{shingling} technique~\cite{leskovec2014mining} is used to convert a long time series into many short time series chunks. Specifically, at each training iteration, we sample a batch of windows with width $W$ at random start point $[1,T-W-1]$ from the dataset and feed into the model. In some cases we have a dataset $\mcX$ of many independent time series with varying length that may share similar features, such as demand forecasting for large number of items. We can model them with a single deep state space model, we build a pool of time series chunks to be randomly sample from, and forecasting is conducted independently for each single time series.

\section{Experiments}
We first compare our proposed model with baselines in public benchmarks, whereby only time series data is provided without additional exogenous variables. In this case, our devised ARD component is switched off and our model degenerates to traditional time series forecasting model. Then we test our model on datasets with rich exogenous variables to show how the proposed ARD network selects relevant exogenous variables and benefit to make accurate forecasts. Note that we refer to our model as DSSMF.

We implement our model by Pytorch on a single RTX 2080Ti GPU. Let dimension of hidden layer of all NNs in generative model and inference model be 100, and dimension of stochastic latent state $\bz$ be 10. We set window size to be one week of time steps for all datasets, and set sample size $K=5$ of SGVB. We use Adam optimizer \cite{kingma2014adam} with learning rate 0.001. Forecasts distribution are estimated by 1000 trials of Monte Carlo sampling. We use Gaussian emission model where mean and standard deviations are parameterized by two-layer NNs following protocol described in Sec. \ref{sec:GM}.

\subsection{Forecasting using Exogenous Variables without Uncertainty}
\subsubsection{Dataset} We use two public datasets \textit{electricity} and \textit{traffic} \cite{yu2016temporal}. The \textit{electricity} contains time series of the electricity usage in $kW$ recorded hourly for 370 clients. The \textit{traffic} contains hourly occupancy rate of 963 car lanes of San Francisco bay area freeways. We held-out the last four weeks for forecasting, and other data for training. For both datasets, only time-feature are added as exogenous variables $\bu_{1:T+\tau}$ without any uncertainty, including absolute time, hour-of-day, day-of-week, is-workday, is-business-hour.

\subsubsection{Baselines} We use a classical method SARIMAX~\cite{box2015time}, extending ARIMA that supports modeling seasonal components and exogenous variable. Optimal parameters are obtained from grid search using R's \textit{forecast} package \cite{hyndman2007automatic}. We also use a recently proposed RNN-based auto-regressive model DeepAR \cite{flunkert2017deepar}, where we use SDK from Amazon Sagemaker platform\footnote{https://docs.aws.amazon.com/sagemaker/latest/dg/deepar.html} to obtain results. Results of all models are computed using a rolling window of forecasting as described in \cite{yu2016temporal}, where the window is one week (168 time steps). 

\subsubsection{Evaluation Metric} Different from deterministic forecasts that predict a single value, probabilistic forecasts estimate a probability distribution. Accuracy metrics such as MAE, RMSE are not applicable to probabilistic forecasts. The \textit{Continuous Ranked Probability Score} (CRPS) generalizes the MAE to evaluate probabilistic forecasts, and is a proper scoring rule~\cite{gneiting2014probabilistic}. Given the true observation value $x$ and the cumulative distribution function (CDF) of its forecasts distribution $F$, the CRPS score is given by:
\begin{equation}
    \text{CRPS}(F, x) = \int_{-\infty}^{\infty}(F(y)-\mathds{1}(y-x))^2\mathrm{d}y
\end{equation}
where $\mathds{1}$ is the Heaviside step function. In order to summarize the CRPS score for all time series with different scales, we compute CRPS score on a series-wise standardized dataset.
\begin{table}[tb!]
	\begin{center}
		\begin{tabular}{l|cccc}
			\toprule
		datasets	& SARIMAX & DeepAR& DSSMF\\
			\hline
			{electricity} & 1.13(0.00) & 0.60(0.01) & \textbf{0.58(0.01)}\\
			{traffic} & 1.58(0.00) & 0.89(0.01) & \textbf{0.86(0.01)}\\
			\hline
		\end{tabular}
	\end{center}
\caption{Mean (S.D.) CRPS for rolling-week forecast over 4 weeks. }
\end{table}\label{tab:benchmark}

The results shown in Table \ref{tab:benchmark} show that our model outperforms DeepAR and SARIMAX on two public benchmarks. 

\subsection{Forecasting using Exogenous Variables\\
with Uncertainty}
In the following, we further show the interpretability of our model by incorporating the exogenous information. The hope is that the relevant exogenous variables will be uncovered automatically by our ARD network which suggests their importance to the forecasting task at hand.

We refer to our model as DSSMF, and we also include two variants for ablation study: 1) DSSMF-NX, DSSMF without using exogenous variable. 2) DSSMF-NA, DSSMF without the proposed ARD network, and the exogenous variable are input directly. To validate if the devised ARD network works effectively, we add three time series of Gaussian white noise as extra exogenous variables (\{\textit{noise 1, noise 2, noise 3}\}). Our expectation is that these three noise variable will be automatically down-weighted by the ARD module. All exogenous variables are standardized to unit Gaussian. To mimic the real situation for forecasting where exogenous variable may be uncertain, we let $u_{d,t}\sim \mcN(\tilde{u}_{d,t},\sigma_t)$ for all exogenous variables except for time-features i.e. exogenous variables without uncertainty in the forecasting period, where $\tilde{u}_{d,t}$ is the true exogenous variable, and $\sigma_t$ increases from 0 to 1 linearly in the forecasting period.

\subsubsection{Electricity Price and Load Forecasting}
We apply our model to two probabilistic forecasting problems using datasets published in GEFCom2014 \cite{hong2016probabilistic}. We choose this benchmark as it contains rich external information to explore by using the devised ARD network. The first is electricity price forecasting, which involves uni-variate time series of hourly zonal electricity price. In addition, zonal load forecasts and total system forecast are also available over the forecast horizon. The second task is electricity load forecasting, where the hourly electricity load and air temperature data from 25 weather stations are provided. 

\subsubsection{Air Quality Forecasting}
We apply our model to the Beijing PM2.5 dataset \cite{liang2015assessing} for the city suffering heavy air pollution at that time. It contains hourly PM2.5 records together with meteorological data, including dew point, temperature, atmospheric pressure, wind direction, window speed, hours of snow, and hours of rain. We aim to forecast PM2.5 given past PM2.5, accurate meteorological data in the training period, and inaccurate meteorological data in forecasting period.
\begin{figure}[!tb]
    \centering
    \subfigure[{electricity price}]
    {\includegraphics[width=0.45\textwidth]{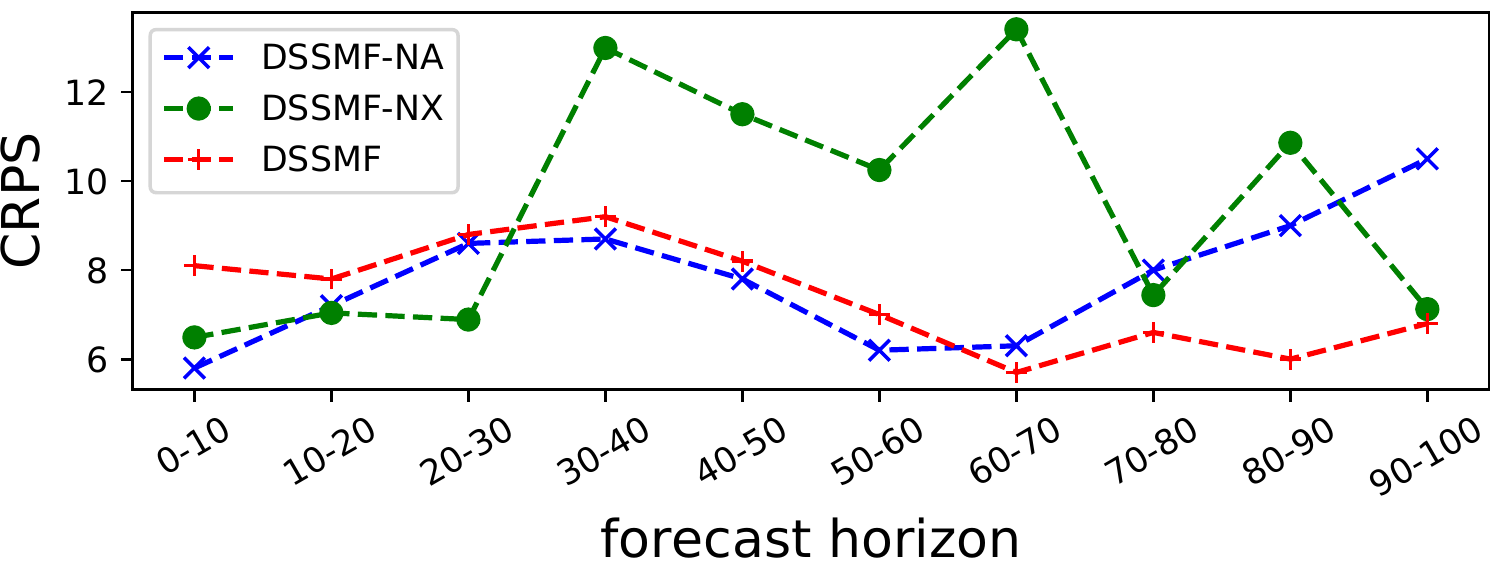}\label{fig:price_horizo}}
    \subfigure[{electricity load}]
    {\includegraphics[width=0.45\textwidth]{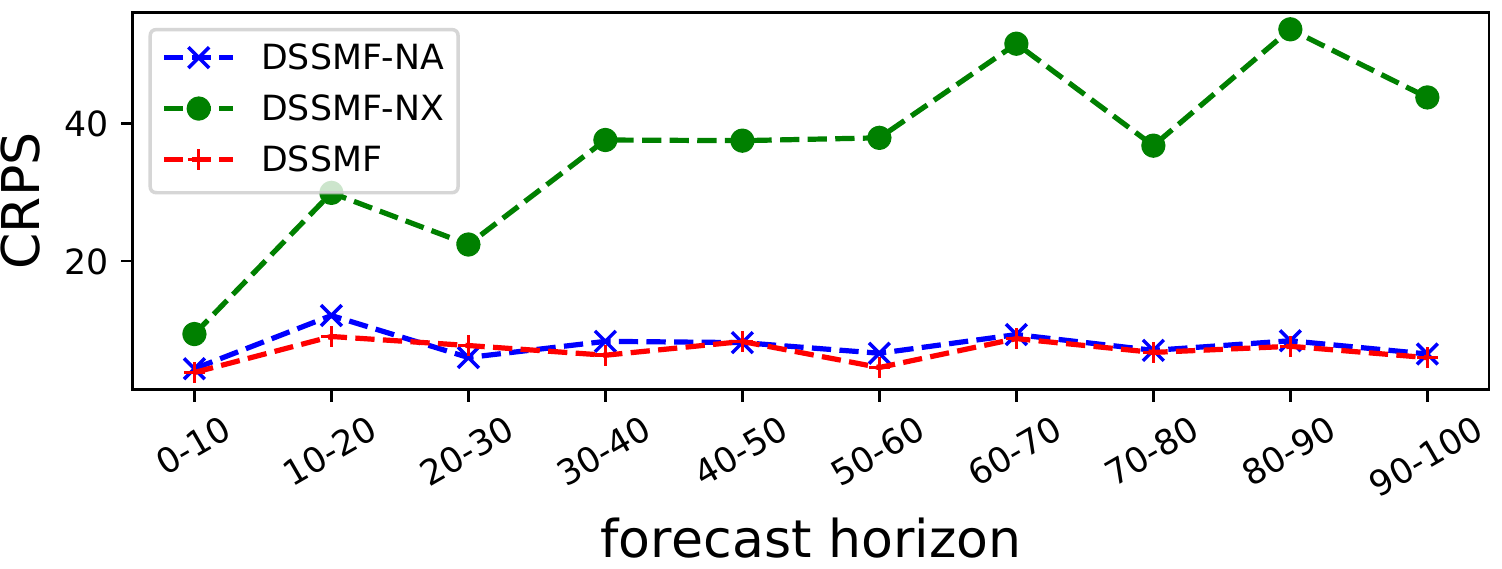}\label{fig:load_horizon}}
    \subfigure[{PM2.5}]
    {\includegraphics[width=0.45\textwidth]{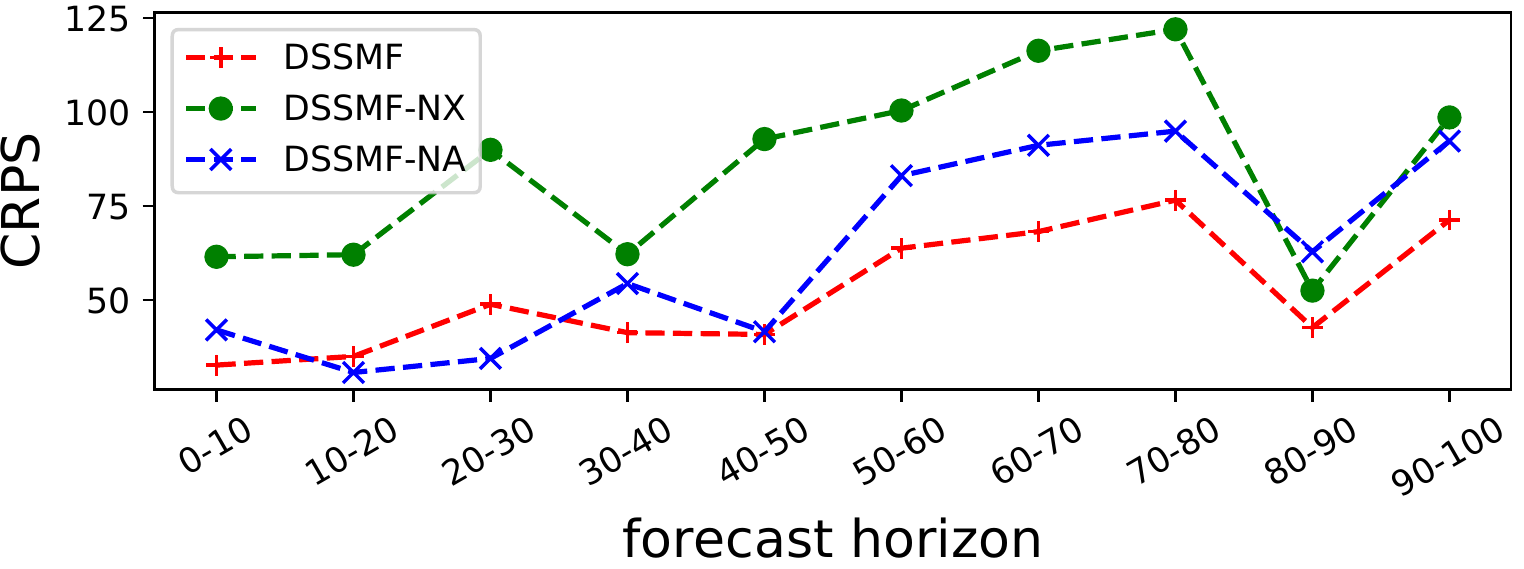}\label{fig:pm25_horizon}}
    \caption{Forecast accuracy w.r.t forecast horizon.}
    \label{fig:horizons}
\end{figure}

\begin{figure}[!tb]
    \centering
    \includegraphics[width=0.48\textwidth]{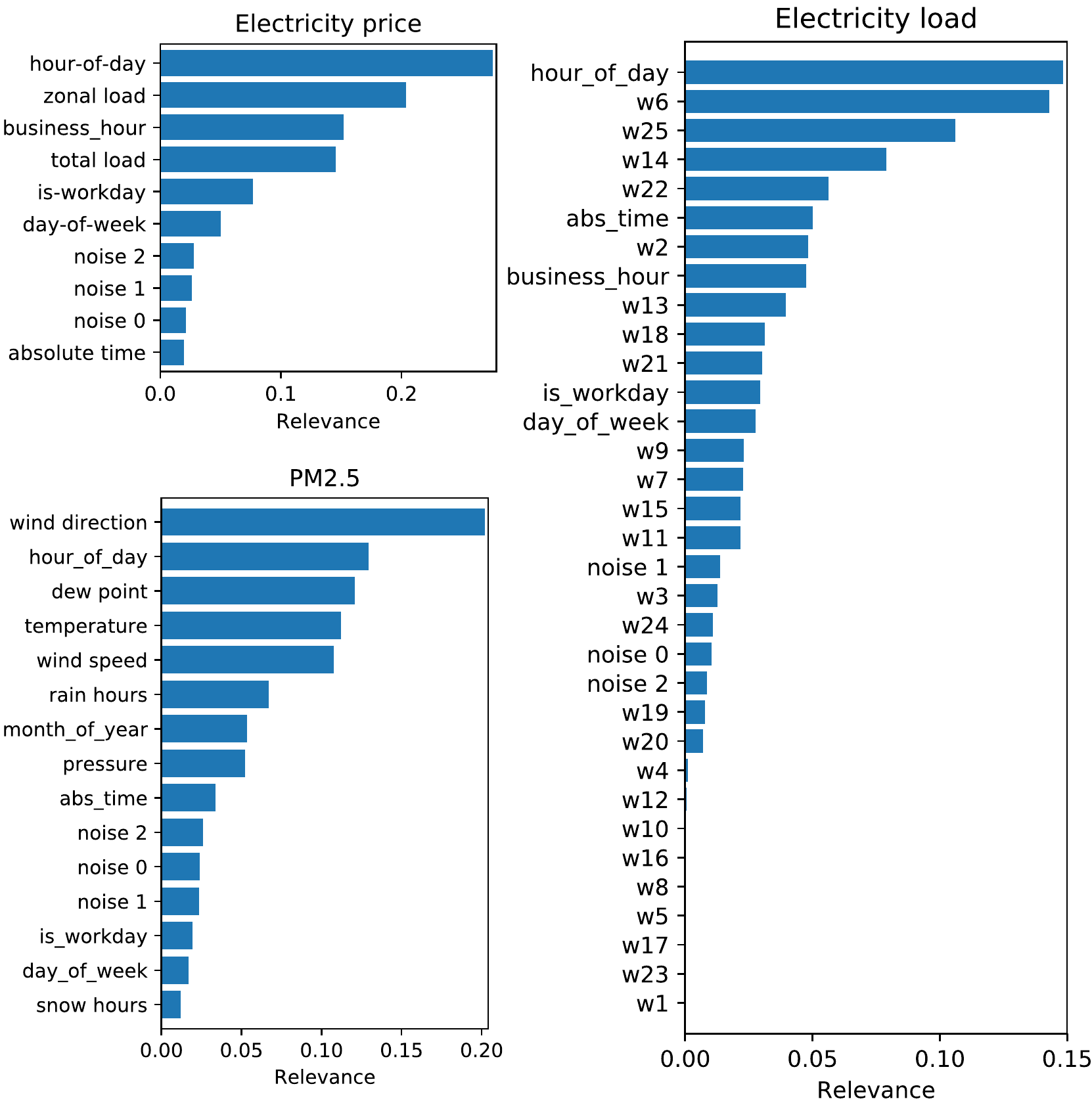}
    \caption{Learned relevance (i.e. $\mathbf{w}$ in Eq.~\ref{eq:relevance_w}) of exogenous variables by the proposed ARD network on three different tasks.}
    \label{fig:ARD_plot}
\end{figure}

Figure~\ref{fig:horizons} shows the forecast accuracy with respect to forecast horizon produced by our main model DSSMF and its two variants. One can see that DSSMF-NX without using exogenous variables performs worst, suggesting that exogenous variables are crucial for making accurate forecasts. Compared with DSSMF-NA which is not equipped with the ARD component, DSSMF performs better in most cases, showing that the ARD network is beneficial for making forecasts.

Figure~\ref{fig:ARD_plot} shows the learned relevance vector $\mathbf{w}$ for the three datasets. It can be found that ARD model has a low relevance weight estimation to the three intentionally added noise  (\{\textit{noise 1, noise 2, noise 3}\}) which are extra exogenous variables. This indicates the effectiveness of our ARD model on identifying irrelevant external factors.

For electricity price forecasting, among its top ranked variables, hour-of-day and business-hour indicate daily seasonality, besides zonal load and total load indicating the demand.

For electricity load forecasting, we use $\{w1,\dots,w25\}$ to denote 25 temperature records from 25 weather stations. Apart from that, it can be found that hour-of-day, absolute time, and business-hour are three most relevant exogenous variables, which also indicate the daily seasonality. For weather data, temperature from the specific station w6, w25, w14 and w22 have the most impact.

For PM2.5 air quality forecasting, we find hour-of-day and month-of-year are two important exogenous variables, which indicates daily and yearly seasonality of PM2.5. Interestingly, wind direction and wind speed are also relevant variables, which indicates that polluted air in Beijing might be blown from other area. Also, this also suggests that wind direction forecast is crucial for making accurate PM2.5 forecasting which in fact has been widely recognized by the public. Besides, meteorological data such as dew point, temperature, and rain are also relevant to PM2.5 forecasting.
\begin{figure*}[!tb]
    \centering
    \includegraphics[width=0.95\textwidth]{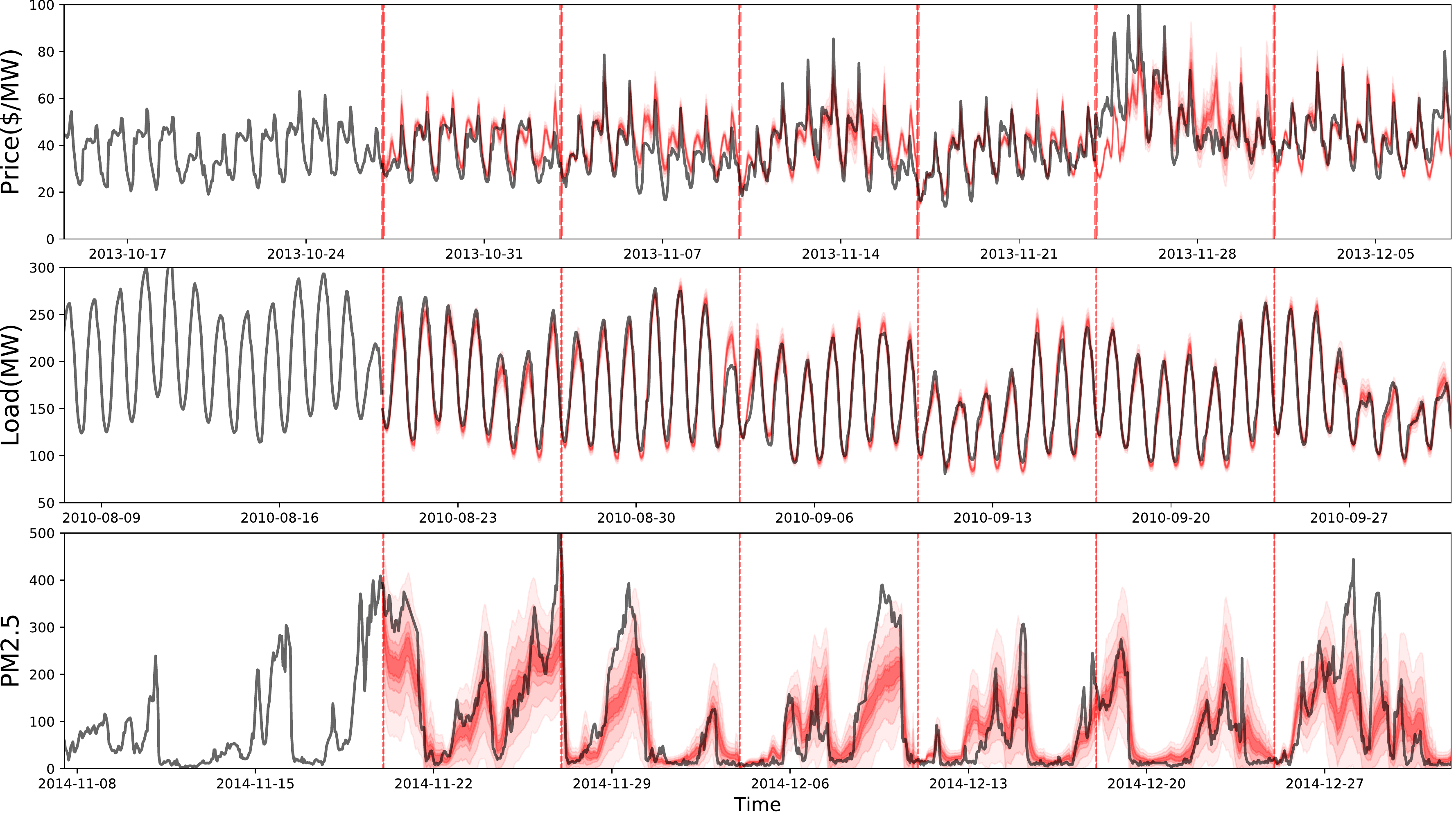}
    \caption{Probabilistic rolling-week multi-step forecasts by our model. Red vertical dashed lines indicate the start time of forecast. Probability densities are plot by confidence intervals [20\%, 30\%,50\%, 80\%, 95\%] in transparency. Past series is not shown in full length. Best viewed in color and zoom in for better view.}
    \label{fig:prob_forecast}
\end{figure*}

Figure \ref{fig:prob_forecast} shows examples for each of the above tasks, we can see that DSSMF makes accurate and sharp probabilistic forecast. For PM2.5 (the third row), the forecast uncertainty (realistically) grows over time (without any extra manipulation on our model). The model spontaneously places higher uncertainty when forecast may not be accurate, which are desired features for downstream decision making process.

\section{Conclusion}
We have presented a probabilistic time series forecasting model, where the future uncertainty is modeled with a generative model. Our approach involves a deep network based embodiment of the state space model, to allow for non-linear emission and transition models design, which is flexible to deal with arbitrary data distribution. Extra exogenous variables in addition to time series data are exploited by our devised automatic relevance determination network, and their uncertainty is considered and modeled by our sampling approach. These techniques help improve the model robustness and interpretability by effectively identifying the key factors, which are verified by extensive experiments.

\clearpage
\newpage
{
\bibliographystyle{named}
\bibliography{main}
}

\end{document}

%% file: commands.tex
\usepackage{amsmath}

\newcommand{\bx}{\mathbf{x}}

\newcommand{\bz}{\mathbf{z}}
\newcommand{\bu}{\mathbf{u}}
\newcommand{\bt}[1][]{\theta_{#1}}
\newcommand{\bphi}{\phi}

\newcommand{\beps}{\bm{\epsilon}}
\newcommand{\bmu}{\bm{\mu}}
\newcommand{\bsigma}{\bm{\sigma}}

\newcommand{\bh}{\mathbf{h}}

\newcommand{\pT}[1][]{p_{\bt[#1]}}
\newcommand{\qPhi}{q_{\bphi}}

\newcommand{\bbR}{\mathbb{R}}
\newcommand{\mcN}{\mathcal{N}}
\newcommand{\mcL}{\mathcal{L}}
\newcommand{\mcX}{\mathcal{X}}